\documentclass[10pt,twocolumn,letterpaper]{article}

\usepackage[preprint]{cvpr}      % To produce the REVIEW version
\usepackage{booktabs}
\usepackage{multirow}
\usepackage{xcolor}
\usepackage{tikz}
\usetikzlibrary{fadings} 
\usepackage[table]{xcolor}
\usepackage{gensymb}
\definecolor{MajutsuBlue}{HTML}{6EB4F9}
\definecolor{MajutsuPink}{HTML}{FF7FAE}
\definecolor{MajutsuPurple}{HTML}{8D88E2}

\definecolor{cvprblue}{rgb}{0.21,0.49,0.74}
\usepackage[pagebackref,breaklinks,colorlinks,allcolors=cvprblue]{hyperref}

\title{
    \textbf{\textit{%
    \textcolor{MajutsuPurple!100!MajutsuPink}{M}%
    \textcolor{MajutsuPurple!90!MajutsuPink}{a}%
    \textcolor{MajutsuPurple!80!MajutsuPink}{j}%
    \textcolor{MajutsuPurple!70!MajutsuPink}{u}%
    \textcolor{MajutsuPurple!60!MajutsuPink}{t}%
    \textcolor{MajutsuPurple!50!MajutsuPink}{s}%
    \textcolor{MajutsuPurple!40!MajutsuPink}{u}%
    \textcolor{MajutsuPurple!30!MajutsuPink}{C}%
    \textcolor{MajutsuPurple!20!MajutsuPink}{i}%
    \textcolor{MajutsuPurple!10!MajutsuPink}{t}%
    \textcolor{MajutsuPurple!0!MajutsuPink}{y}%
    }: Language-driven Aesthetic-adaptive City Generation} \\ 
    \vspace{0.2em}
    \textbf{with Controllable 3D Assets and Layouts} 
}

\author{
Zilong Huang\textsuperscript{\rm 1, *}  \quad Jun He\textsuperscript{\rm 1, *}  \quad Xiaobin Huang\textsuperscript{\rm 1}  \quad Ziyi Xiong\textsuperscript{\rm 1} \\ \quad Yang Luo\textsuperscript{\rm 1}  \quad Junyan Ye\textsuperscript{\rm 1}  \quad Weijia Li\textsuperscript{\rm 1} \quad Yiping Chen\textsuperscript{\rm 1,†} \quad Ting Han\textsuperscript{\rm 1,†} 
    \vspace{5pt} \\
    \textsuperscript{\rm 1} Sun Yat-sen University \\
\textit{    \textsuperscript{\rm *} Equal Contribution  \quad \textsuperscript{\rm †} Corresponding Authors}
}

\begin{document}

\twocolumn
[{
\renewcommand\twocolumn[1][]{#1}%
\maketitle
\begin{center}
\vspace{-0.5cm}
\captionsetup{type=figure}
\includegraphics[width=\linewidth]{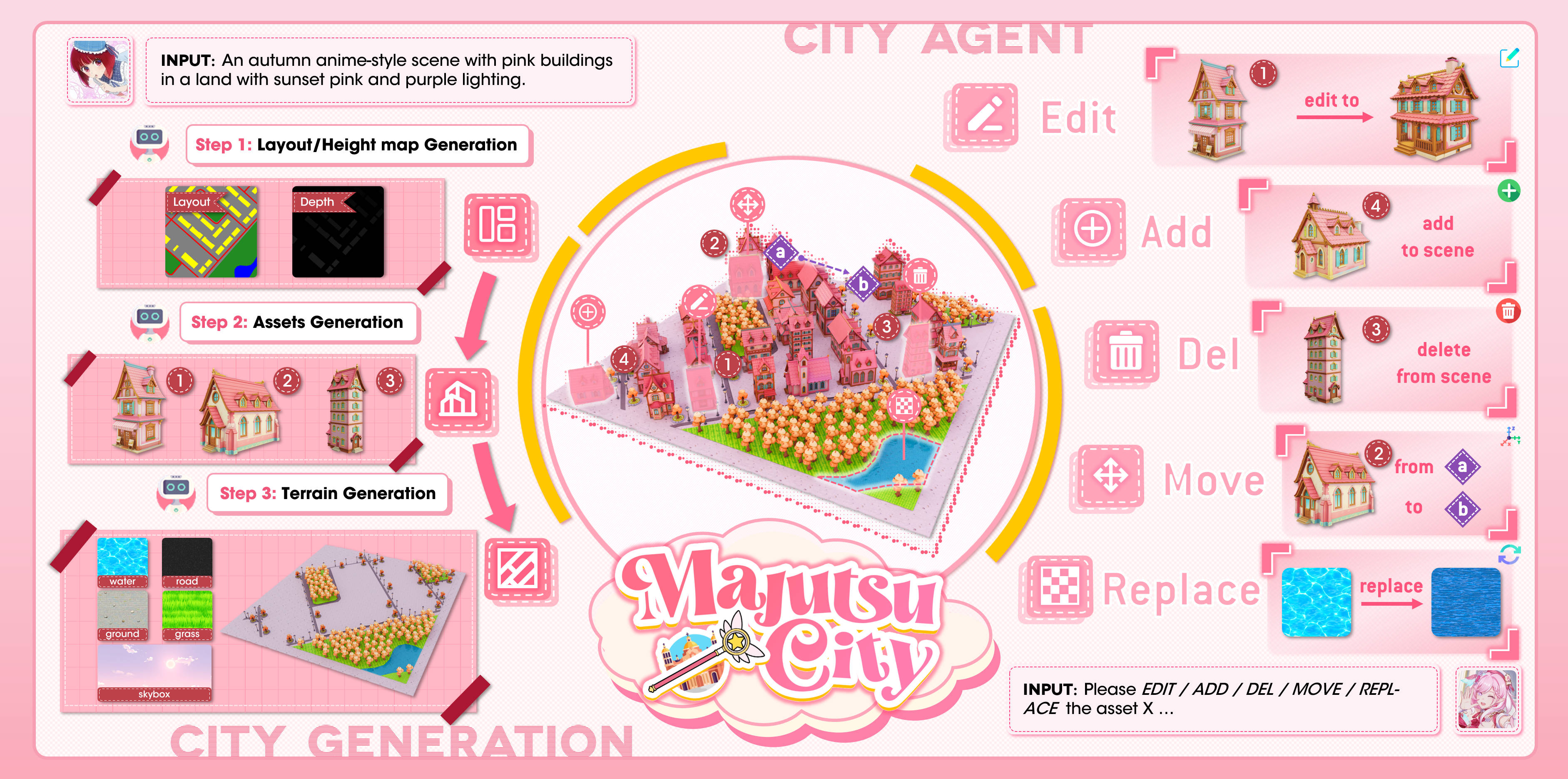}

\captionof{figure}{\textbf{MajutsuCity}
is a language–driven, aesthetic-adaptive system that unifies controllable urban scene generation and interactive editing within a single framework. Conditioned on textual instructions, the framework synthesizes a complete stylized city through layout–height creation, asset instantiation, and terrain/material generation, and further enables iterative refinement through five atomic editing operations. This paradigm forms the core contribution of MajutsuCity, empowering users to create and continuously modify large-scale, stylistically diverse urban scenes through natural language.}

\label{fig:MajutsuCity}
\end{center}
}]

\begin{abstract}

Generating realistic 3D cities is fundamental to world models, virtual reality, and game development, where an ideal urban scene must satisfy both stylistic diversity, fine-grained, and controllability. However, existing methods struggle to balance the creative flexibility offered by text-based generation with the object-level editability enabled by explicit structural representations. 
We introduce \textbf{\textcolor{MajutsuBlue}{MajutsuCity}, a natural language–driven and aesthetically adaptive framework} for synthesizing structurally consistent and stylistically diverse 3D urban scenes. MajutsuCity represents a city as a composition of controllable layouts, assets, and materials, and operates through a four-stage pipeline. To extend controllability beyond initial generation, we further integrate \textbf{\textcolor{MajutsuPink}{MajutsuAgent}, an interactive language-grounded editing agent} that supports five object-level operations. 
To support photorealistic and customizable scene synthesis, we also construct \textbf{\textcolor{MajutsuPurple}{MajutsuDataset}, a high-quality multimodal dataset} containing 2D semantic layouts and height maps, diverse 3D building assets, and curated PBR materials and skyboxes, each accompanied by detailed annotations.
Meanwhile, we develop \textbf{a practical set of evaluation metrics}, covering key dimensions such as structural consistency, scene complexity, material fidelity, and lighting atmosphere.
Extensive experiments demonstrate MajutsuCity reduces layout FID by 83.7\% compared with CityDreamer and by 20.1\% over CityCraft. Our method ranks first across all AQS and RDR scores, outperforming existing methods by a clear margin. These results confirm MajutsuCity as a new state-of-the-art in geometric fidelity, stylistic adaptability, and semantic controllability for 3D city generation. We expect our framework can inspire new avenues of research in 3D city generation. Our project page: \href{blue}{https://longhz140516.github.io/MajutsuCity/}
% Our dataset and code will be released at \href{blue}{https://github.com/LongHZ140516/MajutsuCity}.

\end{abstract}     
\section{Introduction}\label{sec:intro}

The rapid progress of world models and cross-modal generative foundation models has significantly advanced 3D generation at both the object-level \cite{lan2024ln3diff, tang2024lgm, chen20253dtopia, xiang2025structured} and the scene-level \cite{chai2023persistent, chen2023scenedreamer, wang2025vistadream, huang2025midi, wen20253d, huang2025scene4u}. It is crucial for robotics simulation \cite{gonzalez2015procedural, wang2024grutopia, katara2024gen2sim}, virtual reality \cite{thompson2006virtual, nguyen2016applying, jamei2017investigating}, and digital content creation \cite{hendrikx2013procedural, tan2016evolution, li2023omnicity}. City-scale 3D generation remains particularly challenging due to its vast spatial extent, complex topology, and highly diverse architectural styles.

Recent LLM-driven approaches (e.g., SceneCraft \cite{hu2024scenecraft}, 3D-GPT \cite{sun20253d}) offer expressive control but remain limited to small scale, simple scenes, failing to maintain the macroscopic geometric validity required for urban environments. Conversely, layout-guided methods such as InfiniCity \cite{lin2023infinicity}, CityDreamer \cite{xie2024citydreamer}, and GaussianCity \cite{xie2025generative} leverage 2D semantic priors to produce city-level scenes, but rely on implicit or neural rendering representations that suffer from multi-view inconsistency and incompatibility with downstream simulation pipelines.

To bridge the aforementioned gap between representation and application, researchers have turned to exploring urban generation methods based on explicit meshes \cite{xu2024sketch2scene, zhou2024scenex, zhang2024cityx, deng2024citycraft}. Explicit-mesh methods improve structural reliability by retrieving assets from predefined libraries, but their generative diversity is fundamentally constrained by the coverage and style limitations of those libraries, thereby aligning it more closely with a 'Retrieve-and-Place' paradigm than with generation.

To overcome these limitations, we present \textbf{MajutsuCity}, a natural language–driven, aesthetic-adaptive, and fully controllable framework for 3D city generation. Our key insight is that \textit{natural language inherently encodes both macroscopic geometric logic (e.g., “a bustling downtown with towering skyscrapers”) and fine-grained aesthetic intent (e.g., “pink lighting under the sunset”).} MajutsuCity leverages this property by introducing a structured language-to-city specification pipeline through four stages that (1) parses the input text into geometric and aesthetic specifications, (2) generates layout and height maps, (3) synthesizes stylized assets and materials, and (4) assembles them into a coherent, editable 3D city.

We believe that \textit{a practical city generation framework requires not only controllable generation but also efficient editing capabilities.} Beyond the initial generation, we introduce \textbf{MajutsuAgent}, an integrated language-grounded editor that supports object-level Add, Delete, Edit, Move, and Replace operations, enabling the iterative refinement of generated scenes. To support high-quality and customizable synthesis, we further construct \textbf{MajutsuDataset}, a multimodal dataset integrating 2D semantic layouts with building heights, diverse 3D building assets, and production-level PBR textures and skyboxes.

Moreover, we identify that there are no metrics designed for 3D city scenes and thus fail to evaluate structural correctness, material realism, and multi-view consistency. We introduce a VLM-based evaluation framework that provides both absolute scoring (AQS) and relative dimension ranking (RDR) across four essential dimensions: Structural \& View Consistency, Scene Richness \& Complexity, Material \& Texture Fidelity, and Lighting \& Atmosphere.

Extensive experiments demonstrate that MajutsuCity outperforms CityDreamer, GaussianCity, UrbanWorld, and CityCraft across all metrics. Our method achieves 83.7\% FID reduction over CityDreamer and 20.1\% over CityCraft for layout generation, producing significantly sharper and more topologically coherent city structures. Under both AQS and RDR protocols, MajutsuCity ranks \textbf{1st in all eight dimensions}, reflecting superior geometric fidelity, material realism, and aesthetic adaptability across diverse style domains. Our main contributions are as follows:
\begin{enumerate}
    \item We propose \textbf{\textcolor{MajutsuBlue}{MajutsuCity}}, a novel unified natural language-driven framework enabled by a structured text-to-city specification design for controllable and aesthetic-adaptive 3D scene generation with a meaningful evaluation protocol.
    \item We build \textbf{\textcolor{MajutsuPurple}{MajutsuDataset}}, a comprehensive and high-quality dataset that combines thousands of text-aligned layout/height maps, stylistically diverse geometry-constrained 3D assets, and PBR/HDRI materials.
    \item We introduce \textbf{\textcolor{MajutsuPink}{MajutsuAgent}}, a first interactive natural language editing agent that enables fine-grained object-level manipulation of generated scenes.
\end{enumerate}

\section{Related Work}
\label{sec:relatedwork}

\subsection{Layout Generation} 

Compared to general scene layouts \cite{jyothi2019layoutvae, gupta2021layouttransformer}, urban layouts exhibit significantly higher complexity due to the richer semantic categories and irregular geometric topologies present in real-world cities. While vector-based layout generation methods such as BlockPlanner \cite{xu2021blockplanner} and GlobalMapper \cite{he2023globalmapper} provide a structured formulation, they often suffer from limited semantic representation and struggle to model complex, fine-grained spatial patterns. 

In contrast, recent mask-based generation methods \cite{lin2023infinicity, xie2024citydreamer, deng2025citygen, deng2024citycraft} offer better scalability and geometric fidelity, capturing detailed spatial boundaries while maintaining efficient inference. However, these methods typically lack intuitive user control, specifically in expressing high-level design intent through natural language. 

To address this limitation, we train a language-guided urban layout generation model that aligns fine-grained spatial mask synthesis with user-provided textual descriptions, effectively bridging the gap between high-fidelity mask generation and user-intent controllability.

\subsection{Image-to-3D Generation} 

Recent advances in image generation have provided a new direction for 3D content creation, fundamentally reshaping the synthesis of high-quality 3D assets. Early methods were often limited by low-fidelity geometry and low-resolution textures \cite{poole2022dreamfusion, nichol2022point, jun2023shap, muller2023diffrf}, but current 3D generation frameworks increasingly leverage powerful 2D visual priors to improve the consistency of geometry and textures. In particular, many state-of-the-art works employ visual foundation model embeddings such as DINOv2 \cite{oquab2023dinov2} to extract rich semantic and textural representations, coupled with Vision Transformer (ViT) \cite{dosovitskiy2020image} architectures to guide and regularize 3D geometry generation. 

The rapid evolution of a series of SOTA and open-source models (e.g., Trellis \cite{xiang2025structured}, Step1X-3D \cite{li2025step1x}, and Hunyuan3D 2.0 / 2.1 \cite{zhao2025hunyuan3d, hunyuan3d2025hunyuan3d}, as well as commercial tools (e.g., Tripo, Rodin \cite{wang2023rodin}, Meshy, Hunyuan3D 2.5 \cite{lai2025hunyuan3d}, and Hitem3D) have demonstrated that photorealistic and high-detail 3D assets are now not only feasible but increasingly reliable. 

Motivated by these advances, this work explores a principled integration of advanced 3D asset generation models as core components into a city-scale pipeline, enabling large-scale, controllable, and aesthetic-adaptive 3D city generation within a unified, reproducible framework.

\subsection{City Scene Generation} 

In urban scene generation, existing methods such as InfiniCity \cite{lin2023infinicity}, CityDreamer \cite{xie2024citydreamer}, and Persistent Nature \cite{chai2023persistent} have demonstrated the ability to generate large-scale 3D scenes. However, they often rely on implicit or neural representations, resulting in two critical bottlenecks: (1) geometric artifacts and multi-view inconsistency, which stem from the inherent ambiguity of implicit fields, and (2) the absence of explicit, editable object-level structures, making them unsuitable for downstream applications that require precise interaction, editing, and simulation compatibility. 

On the other hand, Procedural Content Generation (PCG)-based techniques can produce highly structured cities \cite{zhang2024cityx, zhou2024scenex, duan2025latticeworld}, but they follow a fundamentally Retrieve-and-Place paradigm. As a result, the diversity and expressiveness of the generated scenes are strictly limited by the scale, style coverage, and quality of the predefined asset libraries, restricting their ability to generalize to novel or stylistically distinctive demands.

Recent advanced works in indoor scene generation have successfully validated a new path: combining the powerful priors of 2D vision models with on-demand 3D object generation to achieve object-level, controllable scene synthesis \cite{dai2024automated, huang2025midi, yao2025cast}. Inspired by this, we aim to scale this object-centric generative paradigm from the indoor level to the macroscopic urban level, enabling a unified framework that is not only controllable and editable, but also adaptive to diverse aesthetic styles, and addressing the core limitations of prior approaches in urban scene generation.
\section{MajutsuCity}

\begin{figure*}[t!]
    \centering
    \setlength{\abovecaptionskip}{0.1cm}
    \setlength{\belowcaptionskip}{-0.2cm}
    \includegraphics[width=1.0\linewidth]{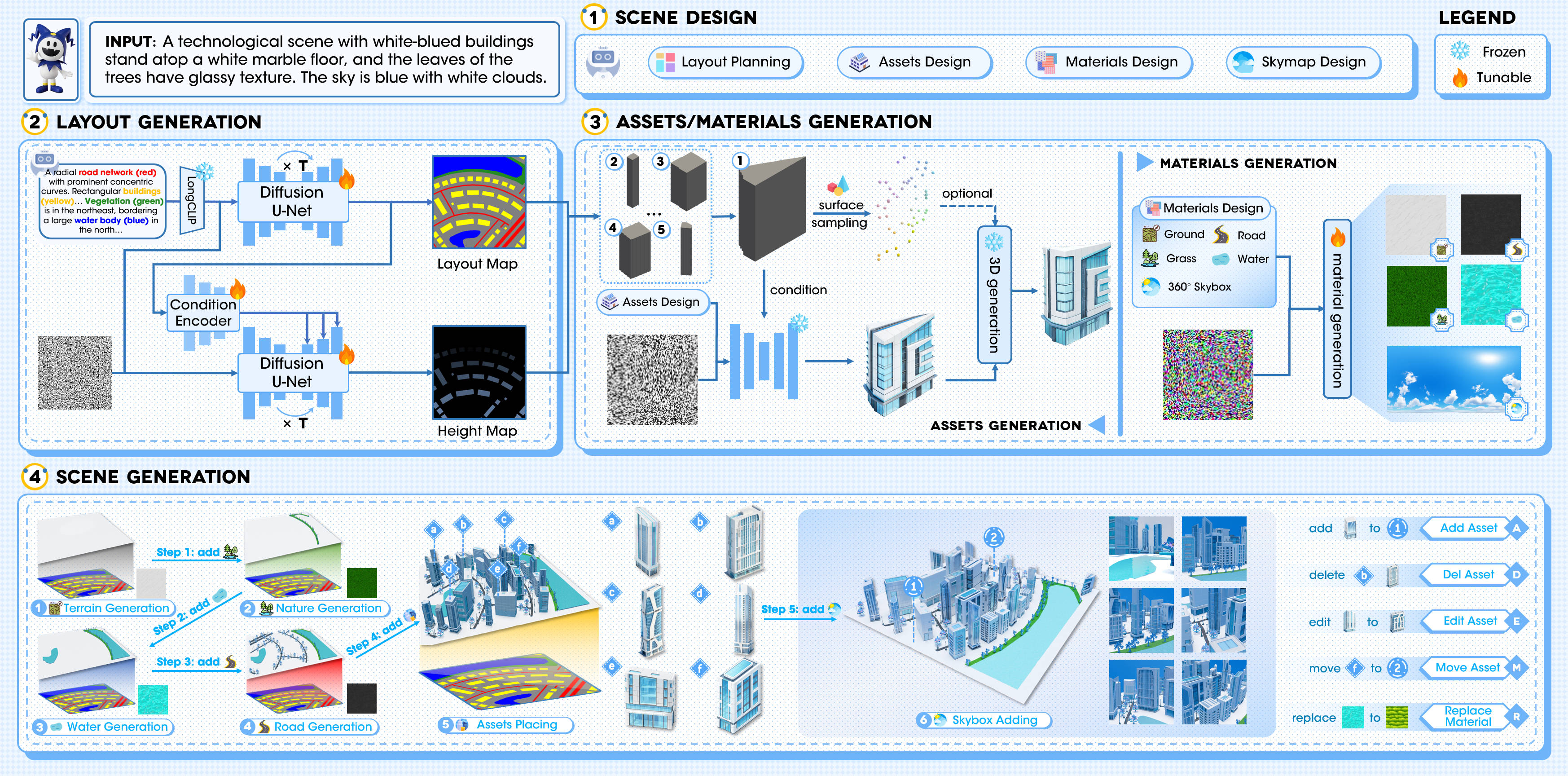}
    \caption{\textbf{Overview of the proposed \textcolor{black}{MajutsuCity} framework.} MajutsuCity is an aesthetic-adaptive generative framework that enables controllable, object-level 3D urban scene generation from natural language descriptions. It consists of Scene Design, Layout Generation, Assets \& Materials Generation, and Scene Generation.}
    \label{fig:pipeline}
    \vspace{-8pt}
\end{figure*}

To enable controllable generation of object-level urban scenes directly from natural language, we propose \textbf{\textcolor{MajutsuBlue}{MajutsuCity}} framework that bridges high-level textual intent and structured 3D scene composition. As shown in Fig.~\ref{fig:pipeline}, the framework has four major stages:
\begin{itemize}
    \item \textbf{Scene Design}: converting textual requirements into structured and consistent design guidance.
    \item \textbf{Layout Generation}: synthesizing spatially coherent urban layouts and height maps under semantic and topological constraints.
    \item \textbf{Assets \& Materials Generation}: producing high-fidelity building-level 3D assets and material maps and skybox.
    \item \textbf{Scene Generation}: composing assets and environmental layers into a coherent and renderable 3D city.
\end{itemize}
Furthermore, we develop MajutsuAgent, an interactive editing agent enables fine-grained, human-in-the-loop scene manipulation with high controllability and consistency.

\subsection{Overview of Scene Design}
\label{sec:scene_design}

Free-form natural language descriptions of urban scenes are typically ambiguous, lacking quantitative and relational constraints. To this end, we employ LLMs for intent understanding and structured decomposition, natural language into structured, executable urban design specifications.

Specifically, the LLM reasons over user prompts to infer latent planning intent and decomposes it into a multi-dimensional urban design template covering Layout, Assets, Materials, and Skymap. Each dimension is parameterized through standardized templates (e.g., land use, spatial distribution, architectural style, and facade material), forming a semantically grounded blueprint that guides both spatial layout generation and 3D asset synthesis.

\subsection{Layout Generation}\label{sec:layout_gen}

Upon receiving the fine-grained spatial guidance $C_{layout}$ from Section~\ref{sec:scene_design}, the goal of this module is to synthesize a spatially aligned pair: a semantic layout map $I_{layout}$ and a building height map $I_{height} $. To address the complex mapping from high-level textual guidance to semantic and geometric outputs, we design a two-stage cascaded framework for layout generation.

The first stage employs a diffusion model $\epsilon_{\theta}^{(1)}$ to synthesize $I_{layout}$ from the detailed long-text description $C_{layout}$. Since $C_{layout}$ (e.g., “a main road stretches from northwest to southeast, flanked by dense commercial buildings”) typically exceeds the token length and contextual limits of the standard CLIP \cite{radford2021learning} encoder, inspired by \cite{ye2025satellite}, we replace the original text encoder $\tau_{\theta}$ with LongCLIP \cite{zhang2024long}. This substitution yields uncompressed, information-rich semantic features $e_c = \tau_{\theta}(C_{layout})$ that provide precise, fine-grained control over complex urban layouts.

{In the second stage, the generated $I_{layout}$ serves as a strong spatial prior $C_s$ (e.g., its semantic mask) and is fed into a ControlNet-based \cite{zhang2023controlnet, li2024crossviewdiff, ye2025skydiffusion} architecture $\epsilon_{\theta}^{(2)}$. Zero-convolution layers inject pixel-level control signals to guide the synthesis of $I_{height}$, ensuring strict spatial consistency with the building regions defined in $I_{layout}$.

Both stages are trained on MajutsuDataset and optimized using the standard latent diffusion objective:
\begin{equation}
\mathcal{L} = \mathbb{E}_{z_0, c, \epsilon \sim \mathcal{N}(0,1), t} \left[ \left\| \epsilon - \epsilon_{\theta}(z_t, t, c) \right\|_2^2 \right].
\label{eq:ldm_loss}
\end{equation}
This decoupled two-stage design enables effective propagation of both high-level semantic intent and low-level spatial constraints, facilitating a coherent generation from textual spatial guidance to layout and elevation map synthesis.

\subsection{Assets and Materials Generation}

\textbf{Asset Generation}. Existing urban scene generation frameworks often exhibit weak coupling between semantic representation and geometric controllability, leading to limited object-level editability and inconsistent structural logic. To this end, we propose a bottom-up asset-level generation paradigm. Given the synthesized layout and building height maps, we extract instance-level building units and generate a dedicated 3D asset for each instance following the Assets Design specification. This formulation decouples global layout generation from local geometric modeling, enabling semantically consistent and structurally controllable asset synthesis. To further ensure spatial alignment and geometric coherence with the prescribed layout, we introduce a shape-constrained generation mechanism composed of two complementary strategies:

\begin{itemize}
    \item \textbf{\textit{Image-based shape constraints}}. Inspired by Qwen-Image-Edit \cite{wu2025qwen}, we introduce an image-guided refinement process that enhances coarse geometries derived from extruding building instance masks in the layout with height-map supervision. Specifically, an isometric rendering $I_{iso}$ of the coarse geometry serves as a geometric prior, while the “Assets Design” prompt $p_{AD}$ provides semantic and stylistic guidance for refining appearance details. This dual conditioning allows the model to enrich visual fidelity without violating the original geometric proportions. To further ensure structural consistency, we incorporate a VLM-based self-calibration mechanism that quantitatively evaluates the shape agreement between the refined result $I_{ref}$ and its prior $I_{iso}$. When a deviation beyond the preset threshold is detected, the system triggers an automatic review–regenerate loop, progressively adjusting the generation parameters until the output satisfies geometric consistency criteria. %This iterative feedback mechanism ensures faithful adherence to both visual semantics and geometric constraints.}
    \item \textbf{\textit{Point cloud-based shape constraints}}. To enhance 3D consistency and ensure strict adherence to the predefined footprint in both shape and height, we introduce an optional point cloud-based constraint strategy. A uniformly sampled point cloud from the coarse geometry serves as an explicit 3D constraint. The sampled point cloud $P_c$, together with a reference image $I_{ref}$ as an appearance guide, is fed into a multi-conditional 3D generation framework \cite{hunyuan3d2025hunyuan3d} to produce the final 3D asset. This joint conditioning guarantees that the synthesized 3D asset remains tightly aligned with the coarse geometry in both shape and scale.
\end{itemize}
In addition to buildings, 3D models of trees and streetlights are also generated to enrich scene diversity.

\textbf{Material Generation.} Unlike discrete building assets, surface features (such as roads, grass, and water surfaces) spatially continuous and require seamlessly tilable textures. Without proper tiling, visible seams and periodic artifacts can occur, degrading overall realism. To overcome the limitations of generic image generators for tilable material synthesis, we adopt Qwen-Image as our visual backbone and fine-tune it on two domain-specific datasets: MajutsuDataset-Material for seamlessly tilable texture maps, and MajutsuDataset-Skybox for high-quality panoramic sky spheres. It produces both material maps and sky spheres, ensuring seamless texture tiling across large-scale urban scenes while improving environmental realism in terms of lighting and atmospheric consistency.

\subsection{Scene Generation}

After completing the Layout Generation and Assets/Materials Generation stages, we integrate all intermediate results to construct a fully renderable, object-level urban scene. The ground, road, water, and vegetation components are organized into four planar layers derived from the semantic layout map, with each layer bound to its corresponding seamlessly tiling material texture to ensure spatial coherence and visual continuity across the scene.

For the vegetation layer, Poisson disk sampling is applied within the vegetation mask to determine tree placement positions, where individual tree models are instantiated. For the road layer, distance-transform-based equidistant sampling is employed to place roadside trees and streetlights along the road boundaries, ensuring realistic spatial distribution and structural consistency.

Subsequently, each building instance in the semantic map is instantiated by placing its shape-constrained 3D asset at the corresponding footprint location after performing a similarity transformation that aligns the local coordinate system to the global layout. Finally, a 360$ \degree $ panoramic sky sphere is integrated into the scene to initialize the ambient lighting and global illumination, yielding a complete, high-fidelity urban scene that remains semantically and geometrically consistent with the layout priors.

\subsection{Scene Agent}

Traditional scene generation pipelines typically lack post-generation editability, whereas object-level scene representations provide a natural interface for fine-grained interaction. Building upon this insight, we develop \textbf{\textcolor{MajutsuPink}{MajutsuAgent}}, a natural language–driven system that enables personalized editing of city-scale scenes. At its core, MajutsuAgent abstracts high-level natural language interactions into five standardized operations, encapsulated within a unified interface:
\begin{itemize}
    \item \textbf{Add:} instantiate and insert new assets into the scene.
    \item \textbf{Delete:} remove specified assets.
    \item \textbf{Edit:} modify the visual or structural attributes of assets.
    \item \textbf{Move:} apply rigid transformations (translation, rotation, scaling) to the selected assets.
    \item \textbf{Replace:} substitute materials on specific surfaces.
\end{itemize}
By leveraging GPT-5 to decompose user commands into a sequence of atomic, interpretable operations, MajutsuAgent accurately translates user intent into controllable scene modifications, thereby enabling intuitive and fine-grained customization of the generated urban environments.
\section{MajutsuDataset}

To enhance the realism and controllability of 3D scene generation, we introduce \textbf{\textcolor{MajutsuPurple}{MajutsuDataset}}, a high-quality multimodal dataset designed for text-guided 3D scene synthesis. As illustrated in Figure \ref{fig:dataset}, the dataset comprises three major components: \textbf{Layout/Elevation}, \textbf{3D Building Models}, and \textbf{Material Assets}. 

\textbf{Layout/Elevation.} Existing urban layout datasets generally lack rich textual annotations that are tightly aligned with visual content, making it difficult for models to learn fine-grained text–layout correspondences and thus constraining controllable generation. To address this challenge, we construct a large-scale city layout dataset based on OpenStreetMap, containing $13,300$ image samples collected from diverse regions representing distinct urban styles across Asia, Europe, South America, Oceania, and America. Each sample consists of a semantic layout map and a building height map, both at a resolution of $512 \times 512$ pixels. The semantic map covers five primary categories (vegetation, roads, water bodies, buildings, and ground). Furthermore, we employ GPT-5-mini \cite{achiam2023gpt} to generate detailed, context-aware textual descriptions for each layout image, enabling fine-grained text-conditional learning.

\textbf{3D Building Models.} With recent advances in 3D generative modeling, it has become feasible to produce high-fidelity, diverse 3D assets on demand. Inspired by this progress, we curate a stylistically diverse library of 3D building models to support downstream urban scene synthesis. Specifically, we define ten representative architectural styles and generate twenty distinct building types for each style. The source images are processed using five commercial-grade 3D generation systems (Meshy, Hunyuan3D, Hyper3D, Tripo3D, and Hitem3D), resulting in a final collection of 1,000 assets with rich stylistic and morphological diversity. All assets will be distributed in full compliance with their respective licensing terms.

\begin{figure}[!t]
    \centering
    \setlength{\abovecaptionskip}{0.1cm}
    \setlength{\belowcaptionskip}{-0.2cm}
    \includegraphics[width=1.0\linewidth]{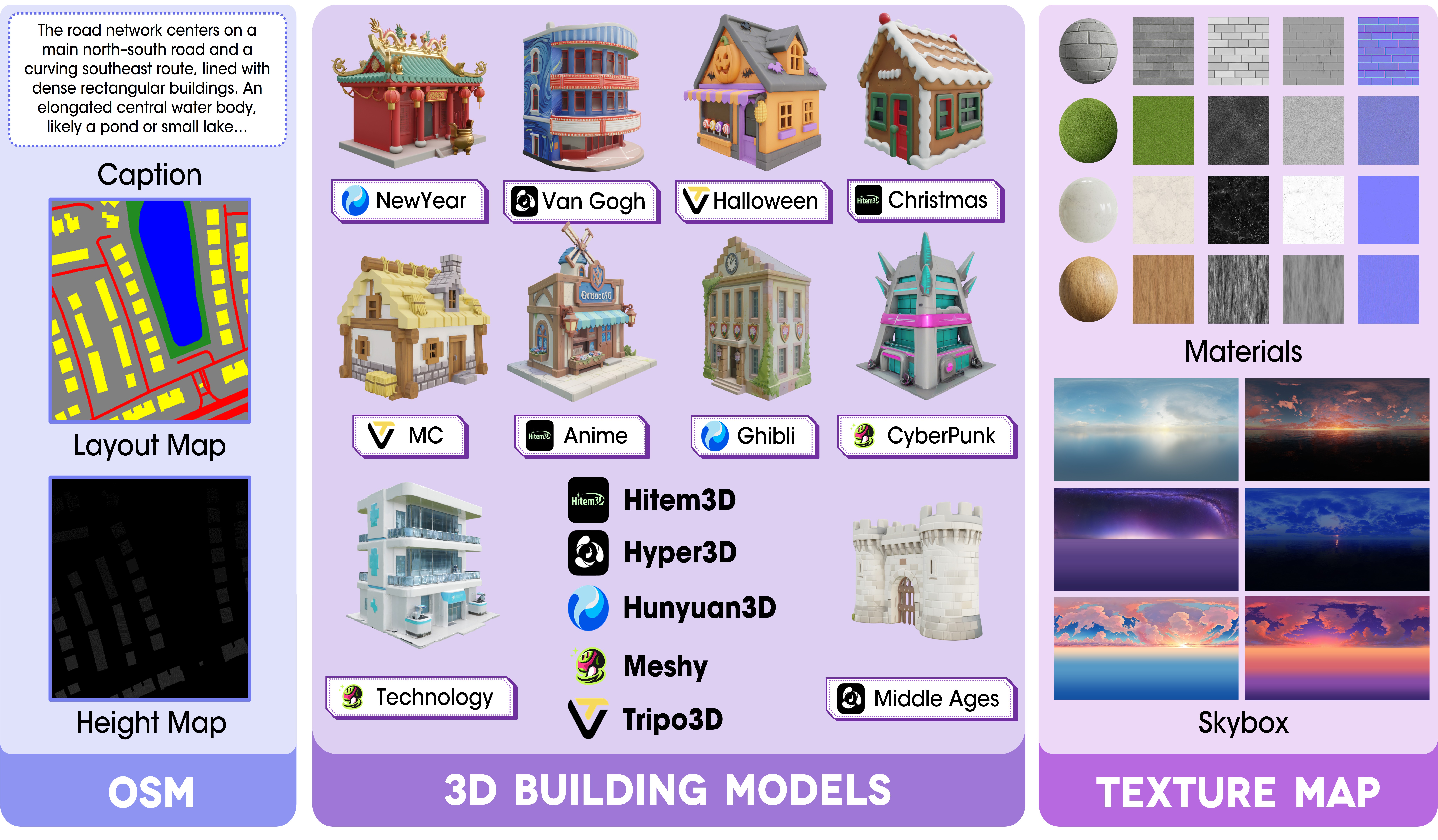}
    % (a)
    \caption{\textbf{Overview of MajutsuDataset}, a high-quality multimodal dataset designed for text-guided 3D urban scene generation. (a) The OSM-based Layout/Elevation subset provides paired semantic layout maps, height maps, and detailed textual descriptions generated by GPT-5-mini. (b) The 3D Building Models subset includes 1,000 assets covering diverse architectural styles. (c) The Texture Map subset contains a large-scale library of seamlessly tilable PBR materials and HDR skybox maps.}
    \label{fig:dataset}
    \vspace{-8pt}
\end{figure}

\textbf{Material Assets.} Existing materials and skybox data often lack physical plausibility, detail fidelity, or seamless tiling capability, limiting their use in production-level rendering. To bridge this gap, MajutsuDataset incorporates a high-quality material asset\footnote{All resources adhere to the CC0 license framework.} library consisting of:
\begin{itemize}
    \item \textbf{PBR Material Library:} $2,300$ seamlessly tilable PBR materials collected from public repositories such as AmbientCG and Poly Haven. Each material includes a complete set of Physically-Based Rendering (PBR) texture maps (Base Color, Normal, Roughness, Metallic, and Ambient Occlusion). 
    \item \textbf{Skybox Map Library:} $1,000$ High Dynamic Range (HDR) skybox maps curated from multiple professional sources, covering diverse lighting conditions and atmospheric contexts.
\end{itemize}
We employ GPT-5-mini to automatically generate detailed textual metadata for each material and skybox asset to enhance semantic usability.
\section{Experiments}

\begin{figure*}[!t]
    \centering
    \setlength{\belowcaptionskip}{-0.2cm}
    \includegraphics[width=1\linewidth]{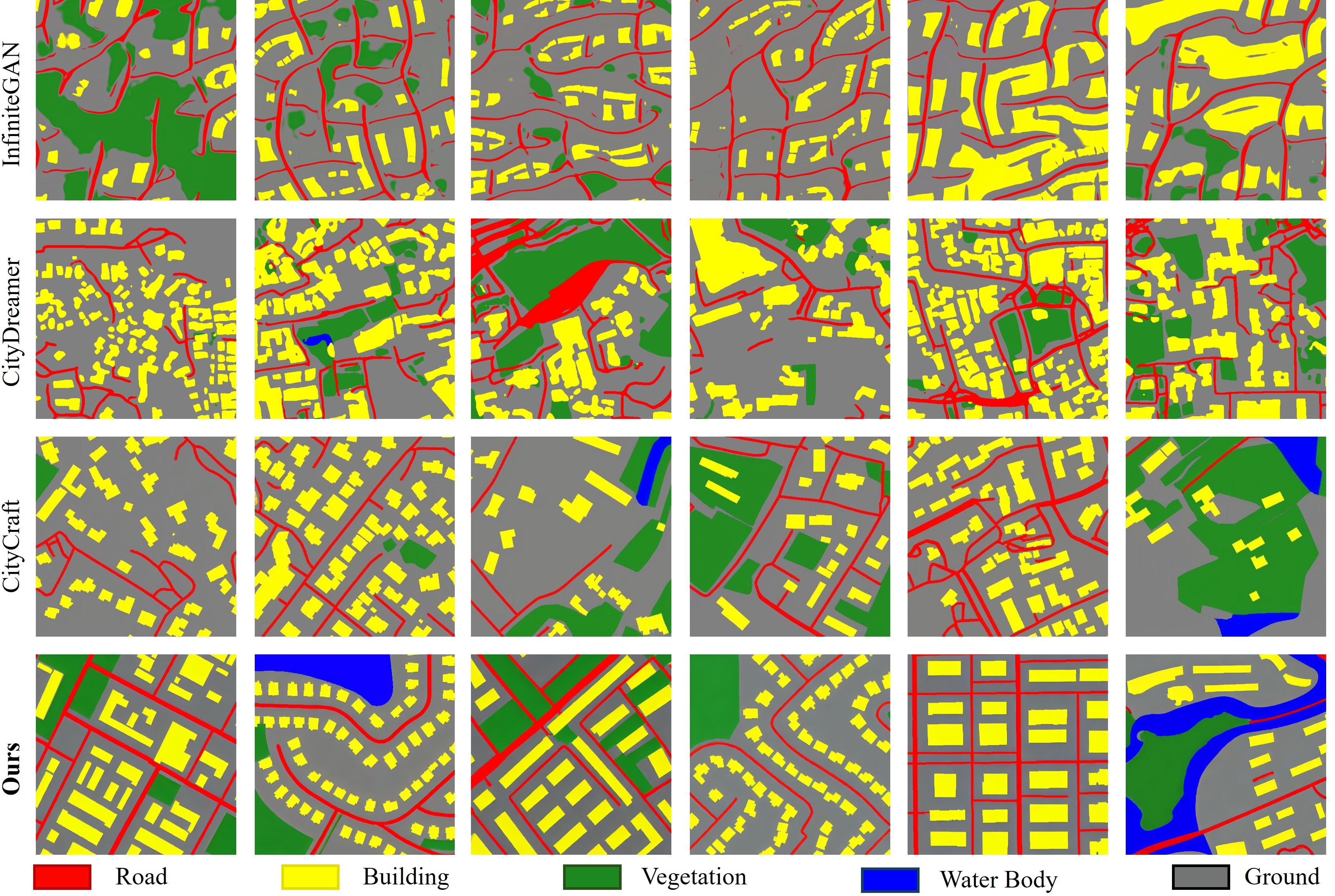}
    \caption{\textbf{Qualitative comparison of city layouts generation}. Our method yields more realistic and coherent urban layouts than prior InfiniteGAN \cite{lin2023infinicity}, CityDreamer \cite{xie2024citydreamer} and CityCraft \cite{deng2024citycraft}.}
    \label{fig:layout}
    \vspace{-8pt}
\end{figure*}

\subsection{Implementation Details and Metrics}

We adopt the pre-trained Stable Diffusion v2.1 model as our baseline for layout generation. The original CLIP text encoder is replaced with LongCLIP. For height map synthesis, a ControlNet encoder is jointly trained with the U-Net to ensure precise spatial alignment between the two stages. Both networks are trained at a resolution of $512 \times 512$ pixels. We employ the AdamW optimizer with an initial learning rate of $1 \times 10^{-5}$. Training is conducted on four NVIDIA A100 GPUs for 100 epochs, using a global batch size of 128. During inference, both stages utilize a Classifier-Free Guidance (CFG) scale of $\omega = 9.0$ and a DDIM sampler with $T = 50$ steps. The entire generation process is executed on a single NVIDIA A800 GPU.

For the City Layout Generation, following the evaluation protocols of CityDreamer \cite{xie2024citydreamer} and CityCraft \cite{deng2024citycraft}, we adopt three widely used metrics to assess the visual fidelity and diversity of the generated city layouts: Fréchet Inception Distance (FID), Kernel Inception Distance (KID), and Inception Score (IS).

For the City Scene Generation, given the absence of a unified and reliable evaluation protocol, we introduce a VLM-based automated evaluation framework motivated by \cite{ye2024loki, zhou2025urbench, fang2024make, ye2025echo}. Our framework is built upon four assessment dimensions: Structural and View Consistency (SVC), Scene Richness and Complexity (SRC), Material and Texture Fidelity (MTF), and Lighting and Atmosphere (LA).

We employ both GPT-5 (as an automated evaluator) and 20 human users in a two-stage procedure: Absolute Quantitative Scoring (AQS) and Relative Dimension Ranking (RDR). In AQS, GPT-5 assigns a 1–10 score for each method based on multi-view renders across the four dimensions, and the mean score is reported as the AQS result.

Inspired by prior work \cite{dubey2016deep, zhang2018measuring, liang2024evaluating, yan2025gpt, he2025urbanfeel, ye2025blink}, we further adopt an RDR protocol to mitigate potential bias inherent in AQS. In RDR, GPT-5 and human users perform pairwise comparisons between sampled image pairs independently for each dimension. Every image participates in at least ten comparisons to ensure robustness. We then aggregate wins and losses per dimension and apply the TrueSkill ranking system \cite{herbrich2006trueskill} to derive dimension-specific RDR scores for all methods.

\begin{table}[!t]
  \fontsize{9pt}{11pt}\selectfont
  \caption{\textbf{Quantitative comparison of Layout Generation.}}
  \label{tab:layout}
  \centering
  \begin{tabular}{@{}l|ccc@{}}
    \toprule
    Method & FID($\downarrow$)  & KID($\downarrow$) & IS($\uparrow$) \\
    \midrule
    InfiniteGAN \cite{lin2023infinicity} & 180.4 & 0.215 & 2.58 \\
    CityDreamer \cite{xie2024citydreamer} & 139.6 & 0.164 & 1.96 \\
    CityCraft \cite{deng2024citycraft} & 28.4 & 0.016 & 3.11 \\
    \midrule
    \rowcolor{MajutsuBlue!10}
    \textbf{Ours} & \textbf{22.7} & \textbf{0.013} & \textbf{3.14} \\
    \bottomrule 
  \end{tabular}
  \vspace{-8pt}
\end{table}

\begin{table*}[t!]
  \caption{\textbf{Absolute Quantitative Scoring (AQS) and Relative Dimension Ranking (RDR) for city scene generation.} 
  For each metric, we report both GPT-based and user-based scores.}
  \label{tab:scene_quantiative_all}
  \centering
  \resizebox{\linewidth}{!}{
  \begin{tabular}{@{}l|cccccccc|cccccccc@{}}
    \toprule
     & \multicolumn{8}{c|}{AQS} & \multicolumn{8}{c}{RDR} \\
    \cmidrule(lr){2-9} \cmidrule(lr){10-17}
    Method & \multicolumn{2}{c}{SVC($\uparrow$)} 
    & \multicolumn{2}{c}{SRC($\uparrow$)} 
    & \multicolumn{2}{c}{MTF($\uparrow$)} 
    & \multicolumn{2}{c|}{LA($\uparrow$)} 
    & \multicolumn{2}{c}{SVC($\uparrow$)} 
    & \multicolumn{2}{c}{SRC($\uparrow$)} 
    & \multicolumn{2}{c}{MTF($\uparrow$)} 
    & \multicolumn{2}{c}{LA($\uparrow$)} \\
    \cmidrule(lr){2-3}\cmidrule(lr){4-5}\cmidrule(lr){6-7}\cmidrule(lr){8-9}
    \cmidrule(lr){10-11}\cmidrule(lr){12-13}\cmidrule(lr){14-15}\cmidrule(lr){16-17}
     & GPT & User & GPT & User & GPT & User & GPT & User 
           & GPT & User & GPT & User & GPT & User & GPT & User \\
    \midrule
    CityDreamer \cite{xie2024citydreamer} 
      & 4.20 & 5.23 & 6.90 & 7.24 & 2.70 & 6.40 & 3.10 & 6.09 
      & 12.39 & 17.98 & 25.70 & 21.44 & 18.50 & 19.99 & 18.25 & 21.55 \\
    GaussianCity \cite{xie2025generative} 
      & \underline{6.73} & \underline{7.63} & \underline{7.17} & \underline{7.37} & 2.83 & 6.54 & 3.33 & 6.23 
      & 23.56 & 20.40 & \underline{27.17} & 21.84 & 18.78 & 21.47 & 18.29 & 22.42 \\
    UrbanWorld \cite{shang2024urbanworld} 
      & 6.17 & 7.16 & 5.40 & 5.99 & 2.14 & 5.41 & 2.80 & 5.17 
      & \underline{25.31} & \underline{23.89} & 16.06 & 14.10 & 12.39 & 14.11 & 13.85 & 14.85 \\
    CityCraft \cite{deng2024citycraft} 
      & 6.00 & 6.97 & 6.11 & 6.74 & \underline{4.22} & \underline{6.64} & \underline{5.00} & \underline{6.23} 
      & 24.44 & 22.79 & 21.33 & \underline{25.42} & \underline{26.60} & \underline{22.99} & \underline{31.12} & \underline{24.62} \\
    \midrule
    \rowcolor{MajutsuBlue!10}
    \textbf{Ours} 
      & \textbf{8.56} & \textbf{8.35} & \textbf{8.33} & \textbf{8.16} & \textbf{7.00} & \textbf{8.03} & \textbf{6.67} & \textbf{7.67} 
      & \textbf{34.13} & \textbf{26.76} & \textbf{28.76} & \textbf{28.97} & \textbf{28.87} & \textbf{25.49} & \textbf{32.68} & \textbf{25.78} \\
    \bottomrule
  \end{tabular}
  }
  \vspace{-8pt}
\end{table*}

\begin{figure*}[!t]
    \centering
    \setlength{\abovecaptionskip}{0.1cm}
    \includegraphics[width=1\linewidth]{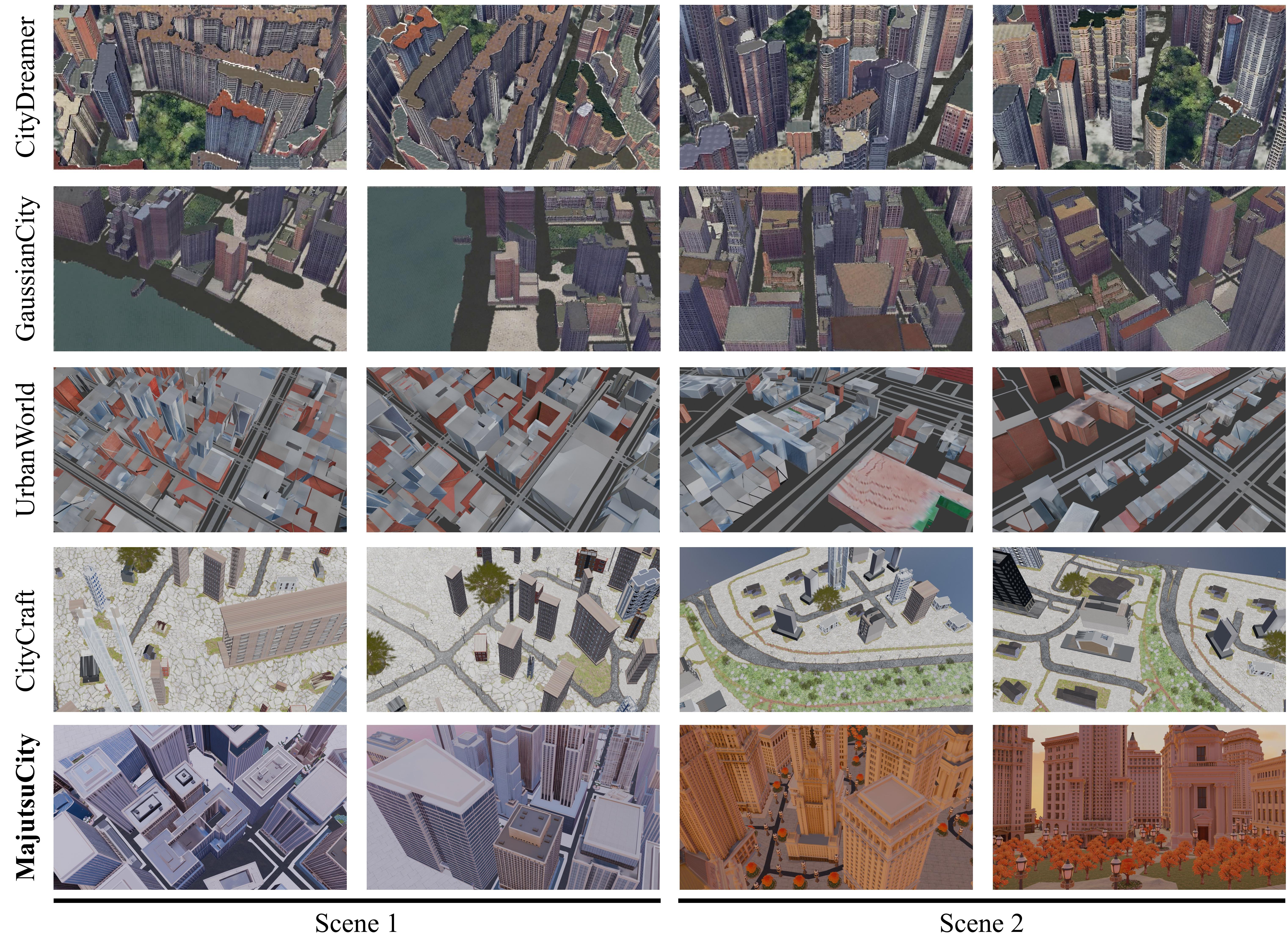}
    \caption{\textbf{Qualitative comparison of city scene}. We compare our method with CityDreamer \cite{xie2024citydreamer}, GaussianCity \cite{xie2025generative}, UrbanWorld \cite{shang2024urbanworld}, and CityCraft \cite{deng2024citycraft} across two representative scenes. Our approach produces scenes with higher geometric fidelity, better multi-view consistency, and richer stylistic diversity than all baselines.}
    \label{fig:scene_qualitative}
    \vspace{-8pt}
\end{figure*}

\subsection{Comparison of City Layout Generation}

\textbf{Qualitative Comparison.} As shown in Figure \ref{fig:layout}, our qualitative comparison with InfiniteGAN \cite{lin2023infinicity}, CityDreamer \cite{xie2024citydreamer}, and CityCraft \cite{deng2024citycraft} reveals that existing methods struggle to preserve structural fidelity. CityDreamer produces fragmented building footprints with poor road continuity, whereas CityCraft, despite generating complete buildings, often yields unrealistic layouts with misaligned structures and implausible spatial distributions. In contrast, our approach generates layouts that are topologically coherent and structurally well-organized. The resulting building footprints are complete, sharp, and consistently aligned with the road network, forming realistic city blocks that provide a reliable structural foundation for high-fidelity downstream scene generation.

\textbf{Quantitative Comparison.} The quantitative results are shown in Table \ref{tab:layout}. Our method achieves improvements of 83.7\% and 20.1\% in FID over CityDreamer \cite{xie2024citydreamer} and CityCraft \cite{deng2024citycraft}, respectively. Benefiting from the guidance of fine-grained spatial text, our model produces layouts that are structurally more plausible and better aligned with the real-world distribution. In addition, the higher IS score demonstrates that our method generates clearer and more diverse urban structures, further validating its effectiveness in large-scale city layout synthesis.

\subsection{Comparison of City Scene Generation}

To assess the quality of the generated city scenes, we compare our method with four state-of-the-art approaches in urban scene synthesis: CityDreamer \cite{xie2024citydreamer}, GaussianCity \cite{xie2025generative}, UrbanWorld \cite{shang2024urbanworld}, and CityCraft \cite{deng2024citycraft}. All 3D assets in our case scenarios are generated by using Hunyuan3D.

\textbf{Qualitative Comparison.} Figure \ref{fig:scene_qualitative} provides two representative examples comparison. CityDreamer and GaussianCity exhibit severe geometric artifacts and multi-view inconsistency, due to limitations of their underlying scene representations. UrbanWorld employs explicit meshes but is restricted to coarse primitives and suffers from low texture fidelity. CityCraft produces visually realistic results but remains highly constrained in stylistic diversity by its fixed asset library. In contrast, our method achieves both stable geometry and strong multi-view consistency, while also delivering rich stylistic diversity and high-fidelity geometry and textures, substantially outperforming all baseline methods in overall visual quality.

\textbf{Quantitative Comparison.} The AQS and RDR results reported in Table \ref{tab:scene_quantiative_all}, clearly demonstrate that our model surpasses all baselines. CityDreamer performs the worst on SVC due to intrinsic limitations of its NeRF-based representation. CityCraft achieves relatively strong performance in MTF and LA, attributed to its high-quality asset library, but its stylistic monotony results in low SRC scores. Importantly, we observe a notable discrepancy that although the 3DGS-based GaussianCity outperforms mesh-based UrbanWorld and CityCraft in SVC under AQS, the ranking is reversed under RDR. This reversal indicates that RDR effectively mitigates the bias present in absolute scoring and provides a more robust assessment of structural consistency. Significantly, the strong alignment between GPT and human evaluations validates our assessment dimensions and confirms GPT's near-human assessment capabilities.

\subsection{Ablation Study}

\textbf{Effectiveness of Layout Generation.} We conduct an ablation study on the layout generation module in Table \ref{tab:layout_ablation} to validate the effectiveness of fine-grained spatial text and the LongCLIP encoder. The results show that removing LongCLIP causes the FID metric to degrade from 22.7 to 28.0, confirming its important role in understanding long-text spatial relationships. Similarly, removing the spatial text guidance (and using short prompts instead) leads to a significant degradation in FID, which rises to 35.7. The KID and IS metrics show consistent trends, indicating that both components are indispensable for generating high-quality, topologically coherent city layouts.

\begin{table}[!t]
  \fontsize{10pt}{11pt}\selectfont
  \caption{\textbf{Ablation Study of Layout Generation.} 'Spatial Text' represents fine-grained spatial text, and 'LongCLIP' represents the long-text visual-language pre-training module \cite{zhang2024long}.}
  \label{tab:layout_ablation}
  \centering
  \begin{tabular}{@{}cc|ccc@{}}
    \toprule
    Spatial Text & LongCLIP & FID($\downarrow$)  & KID($\downarrow$) & IS($\uparrow$) \\
    \midrule
     & & 35.7 & 0.025 & 3.08 \\
     \checkmark & & 28.0 & 0.023 & 3.07 \\
    \rowcolor{MajutsuBlue!10}
     \checkmark & \checkmark & \textbf{22.7} & \textbf{0.013} & \textbf{3.14} \\
    \bottomrule 
  \end{tabular}
  \vspace{-8pt}
\end{table}

\noindent\textbf{Aesthetic-Adaptive Style Transformation.} To further demonstrate the ability of our model to generate stylistically diverse urban scenes, we provide additional qualitative results in Figure \ref{fig:4styles}. We condition the generation on four widely recognized and visually distinctive styles (e.g., Minecraft, Netherlands, CyberPunk, and Ghibli). Our approach not only faithfully captures the defining aesthetic characteristics of each style but also preserves strong intra-style coherence across large-scale city scenes.

\begin{figure}[!t]
    \centering
    \includegraphics[width=1.0\linewidth]{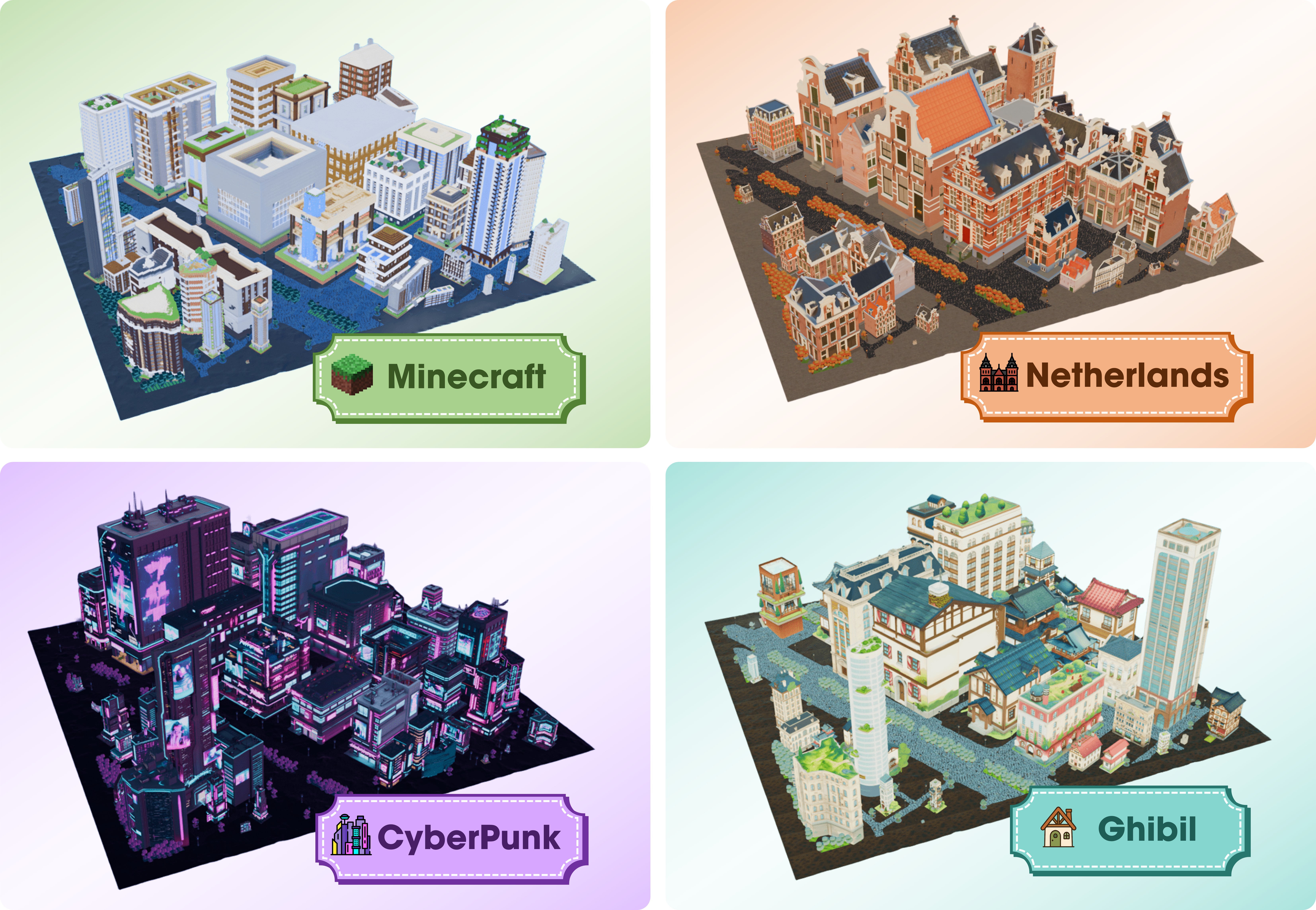}
    \caption{\textbf{Style-driven city generation results.} Four city scenes with different well-known styles generated by MajutsuCity show high fidelity and strong intra-style consistency.}
    \label{fig:4styles}
    \vspace{-8pt}
\end{figure}

% \textbf{Limitations.} Our method may produce suboptimal layouts from text with logical conflicts and geometric inaccuracies when generating assets for buildings with highly irregular structural details.

\section{Conclusion}

In this paper, we introduce MajutsuCity, a natural language–driven and aesthetic-adaptive framework for controllable 3D urban scene generation. Through a unified four-stage pipeline, MajutsuCity enables precise structural control, and supports diverse and customizable visual styles, which remain limited in existing approaches. To establish a complete system, we further develop three key components: (1) MajutsuDataset, a high-quality multimodal dataset that integrates 2D layout/elevation priors, high-fidelity 3D assets, and PBR materials, providing essential data foundations for photorealistic and immersive scene construction. (2) MajutsuAgent, an interactive agent that extends the role of language guidance from initial scene synthesis to post-generation editing. (3) Metric suite consisting of Absolute Quantitative Scoring (AQS) and Relative Dimension Ranking (RDR) systematically assesses structural consistency, scene richness, material fidelity, and lighting realism, offering the first holistic metric system for benchmarking urban scene generation quality. Our work opens new avenues for scalable, controllable, and style-adaptive 3D urban generation. We believe MajutsuCity provides not only a powerful generative framework but also a foundational direction, a powerful dataset, and a meaningful metric for future research in VLM-Based 3D content generation. We will continue to expand and refine the dataset to foster broader research and contribute more substantially to the community.

\section{Limitations and Futurk Work}
\label{furturework}
Despite the significant advancements that MajutsuCity demonstrates in interactive city generation, several limitations remain inherent to the current framework:

\begin{itemize}
    \item \textbf{Sensitivity to Prompt Logical Consistency.} The layout generation module is highly sensitive to the logical coherence of user-provided prompts. Due to the strictly hierarchical nature of our pipeline, contradictory spatial instructions or geometrically infeasible layouts generated during the initial scene design stage can propagate through subsequent modules. These cascading errors may lead to implausible road networks or unreasonable building configurations, ultimately degrading the quality of the assembled scene.
    \item \textbf{Uncontrollability of highly complex shapes.} Although both image-based and point cloud-based constraints are incorporated to preserve geometric fidelity in building generation, ensuring high visual fidelity while maintaining geometric validity remains challenging, particularly for buildings with highly complex or irregular architectural topologies.
    \item \textbf{Visual Scale Inconsistency.} Since each building asset is synthesized independently in its own local coordinate space, the model lacks a global sense of scale and contextual awareness. Consequently, even though our shape constraint strategies effectively regulate individual instance shapes, inconsistent visual scales may occur during final assembly (e.g., inconsistent story heights between adjacent buildings).
\end{itemize}

Recently, Meta introduced WorldGen~\cite{wang2025worldgen}, a system with goals closely aligned with ours: generating fully interactive 3D worlds directly from text prompts. WorldGen adopts a top-down paradigm, “\textit{Holistic Planning} $\to$ \textit{Holistic Scene Generation} $\to$ \textit{Object Decomposition} $\to$ \textit{Object Refinement}”. It first generates a foundationally coherent layout and RGB sketch guided by user instructions, then decomposes the scene into interactive and independent assets using object-aware techniques, followed by geometric and textural refinements. This \textit{Holistic-then-Local} workflow effectively circumvents the scale fragmentation commonly caused by instance-level stitching, while ensuring continuous terrain topology and strong visual stylistic unity.

Inspired by this direction, our future work will explore a hybrid generation paradigm. Our aim is to further enhance the physical rationality and visual harmony of complex urban scenes while maintaining the high level of controllability offered by our current method.

% \clearpage
% \newpage
{
    \small
    \bibliographystyle{ieeenat_fullname}
    \bibliography{main}
}
\end{document}